\pdfoutput=1

\documentclass[11pt]{article}

\usepackage[]{acl}

\usepackage{times}
\usepackage{latexsym}
\usepackage{graphicx}
\usepackage[T1]{fontenc}

\usepackage[utf8]{inputenc}

\usepackage{microtype}

\usepackage{inconsolata}

\usepackage{tcolorbox}
\usepackage{amssymb}
\usepackage{multirow}
\usepackage{enumitem}
\usepackage[bottom]{footmisc}
\usepackage{xcolor,soul}
\sethlcolor{lightgray}

\newtcolorbox{mybox}[1]{
    title={#1},
    colback=white,
    colframe=black,
    fonttitle=\bfseries,
    sharp corners,
    boxrule=0.5pt,
    coltitle=black,
    toptitle=1mm,
    bottomtitle=1mm,
    attach boxed title to top left={yshift=-\tcboxedtitleheight/2},
    boxed title style={
        colback=white,
        colframe=white,
        size=small,
        sharp corners
    }
}

\NewDocumentCommand{\yi}
{ mO{} }{\textcolor{magenta}{\textsuperscript{\textit{Yi}}\textsf{\textbf{\small[#1]}}}}

\NewDocumentCommand{\heng}
{ mO{} }{\textcolor{red}{\textsuperscript{\textit{Heng}}\textsf{\textbf{\small[#1]}}}}

\usepackage{xcolor,colortbl}
\definecolor{green}{rgb}{0.1,0.1,0.1}

\usepackage{xspace}
\newcommand{\datasetname}{\textbf{CultureAtlas}\xspace}

\title{\raisebox{0mm}{\includegraphics[trim=0 24 0 0, width=0.8cm]{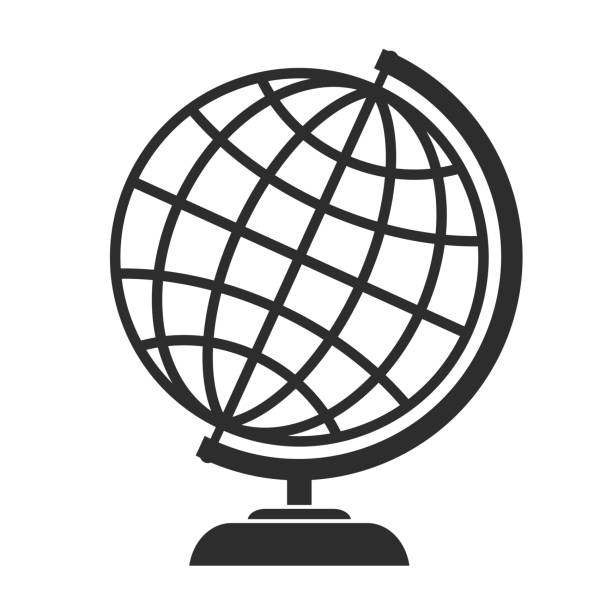}}No Culture Left Behind: \\ Massively Multi-Cultural Knowledge Acquisition \& LM Benchmarking \\ on 1000\textsuperscript{+} Sub-Country Regions and 2000\textsuperscript{+} Ethnolinguistic Groups 
}

\author{Yi R. Fung ~~~ Ruining Zhao ~~~Jae Doo ~~~Chenkai Sun ~~~ Heng Ji \\
  University of Illinois at Urbana-Champaign \\
  \texttt{\{yifung2,hengji\}@illinois.edu} \\ 
  }

\begin{document}
\maketitle

\begin{abstract}
Pretrained large language models have revolutionized many applications but still face challenges related to cultural bias and a lack of cultural commonsense knowledge crucial for guiding cross-culture communication and interactions. Recognizing the shortcomings of existing methods in capturing the diverse and rich cultures across the world, this paper introduces a novel approach for massively multicultural knowledge acquisition. Specifically, our method strategically navigates from densely informative Wikipedia documents on cultural topics to an extensive network of linked pages.  
Leveraging this valuable source of data collection, we construct the \datasetname dataset, which covers a wide range of sub-country level geographical regions and ethnolinguistic groups, with data cleaning and preprocessing to ensure textual assertion sentence self-containment, as well as fine-grained cultural profile information extraction. 
Our dataset not only facilitates the evaluation of language model performance in culturally diverse contexts but also serves as a foundational tool for the development of culturally sensitive and aware language models. Our work marks an important step towards deeper understanding and bridging the gaps of cultural disparities in AI, to promote a more inclusive and balanced representation of global cultures in the digital domain.\footnote{Our code will be released at \url{https://github.com/yrf1/LLM-MassiveMulticultureNormsKnowledge-NCLB}.} 
\end{abstract}
\section{Introduction}

Pretrained large language models (LMs) are increasingly used in applications across diverse domains, ranging from question answering \cite{brown2020language,gangi-reddy-etal-2022-zero} and chatbots \cite{lin2020caire,ouyang2022training} to content recommendation \cite{wu-etal-2020-mind} and norm violation detection \cite{fung2022normsage}. However, as their usage proliferates, an important concern emerges -- the potential for cultural bias and misappropriation \cite{hershcovich-etal-2022-challenges,palta-rudinger-2023-fork,li-etal-2023-defining}. LMs, when not equipped with geo-diverse knowledge and cultural sensitivity, can exhibit significant disparities in performance when faced with textual data from different regions. This imbalance disadvantages certain users groups and exacerbates existing biases in NLP applications, perpetuating the presently predominant Western-centric perspectives and knowledge bases. Hence, tackling this issue is paramount not only for promoting fairness and inclusivity, but also for fostering a more culturally aware and harmonious digital landscape.

\begin{figure}[!t]
\includegraphics[width=7.7 cm]{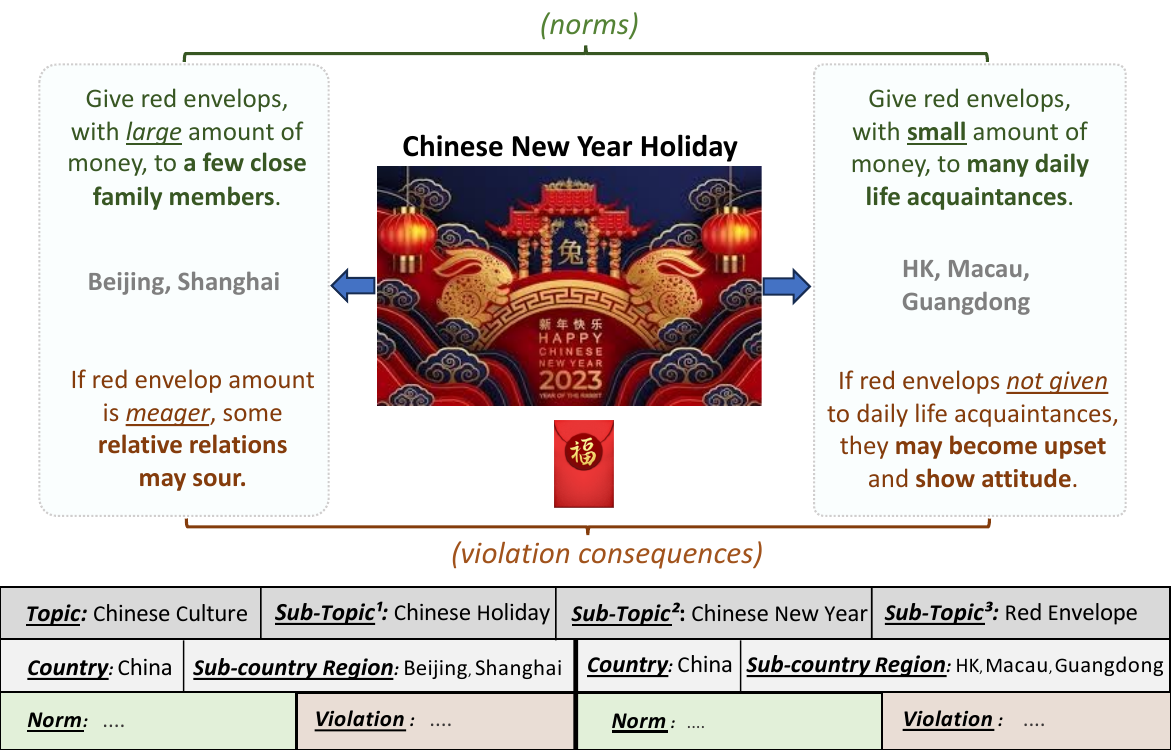}
\centering
\caption{An example illustration of different cultural practices, with regards to Chinese New Year red envelopes across different geographical regions such as \textcolor{gray}{\textit{Beijing/Shanghai}} versus \textcolor{gray}{\textit{HK/Macau/Guangdong}}.}
\label{fig:sugar_candy}
\vspace{-0.9em}
\end{figure}

\begin{figure*}[!t]
\includegraphics[width=15.8cm,trim={0.1cm 0.9cm 0.9cm 1.7cm},clip]{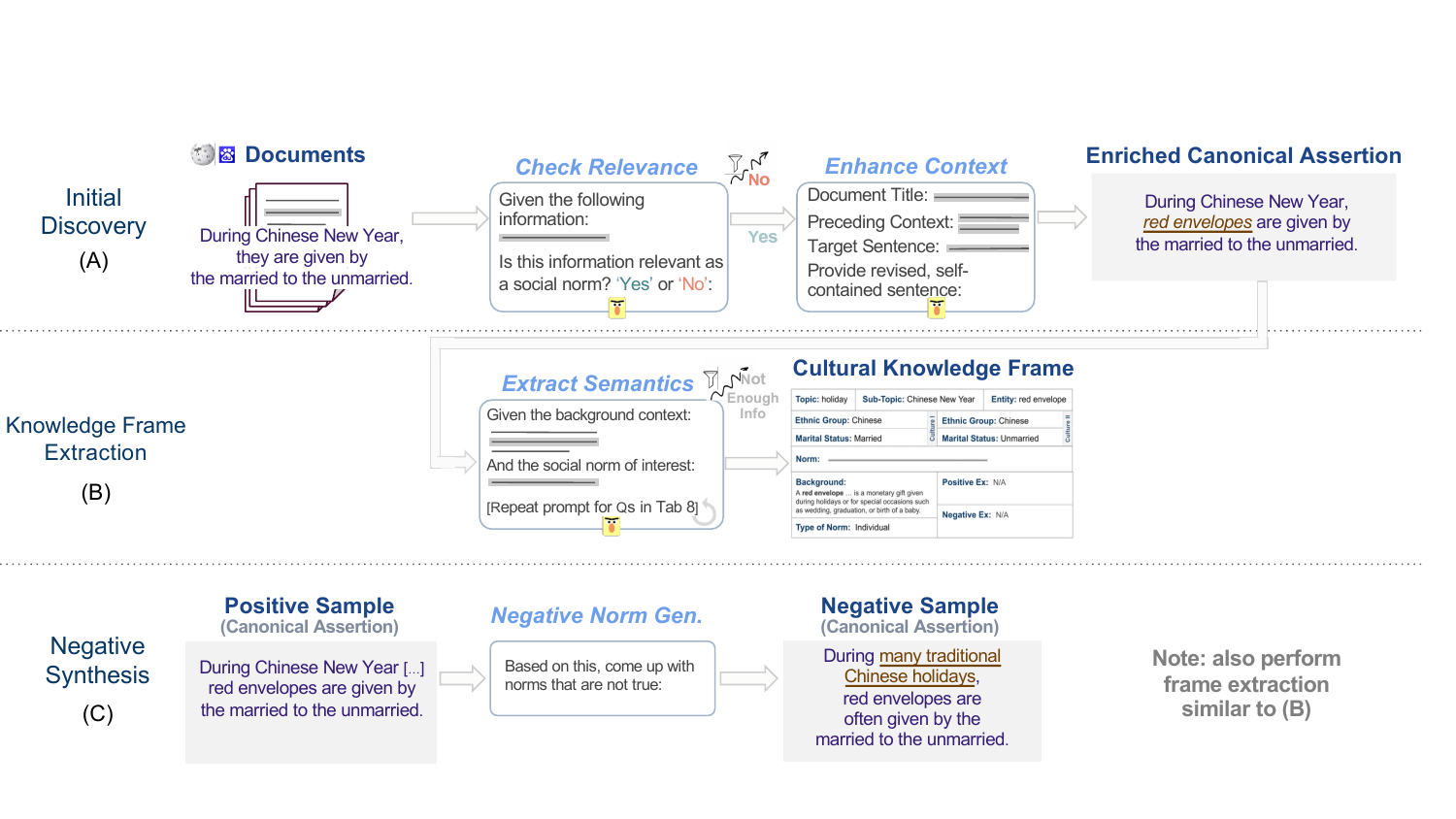}
\centering
\caption{An overarching view of our \datasetname benchmark construction process.}
\label{fig:framework}
\vspace{-0.7em}
\end{figure*}
Previous approaches in benchmarking and improving the cross-cultural knowledge of language models tend to either ~1) focus on a predefined, narrow set of coarse-grained cultures and cultural topics \cite{yin-etal-2022-geomlama}, or ~2) discover cultural knowledge from large noisy corpora \cite{10.1145/3543507.3583535} in which important cultural elements often get filtered out in the data processing stage or lost as cultural differences get intermingled, leading to a scenario where LMs fail to learn information specific to individual subregions. Our goal is to investigate and set the grounds for empowering language model with reasoning capability on finer-grained cultural nuances that pertain to different cultural subgroups and deeper topic coverage. For example, as depicted in Figure \ref{fig:sugar_candy}, Chinese New Year red envelop gifting practices vary by geographical subregions within a same country, which may potentially lead to norm violations for people newly settling into an area who are unaware of the common cultural practice that differ by region or recipient group. 

In this work, we propose a culture knowledge acquisition process for constructing a novel benchmark for assessing language models'  massively multicultural reasoning capabilities. In particular, our culture knowledge acquisition process seeks to combine the best of both worlds between \textit{bottom-up discovery} of culture knowledge discovery from the open web documents (relatively noisy but large-scale data) and \textit{top-down discovery} of culture knowledge from targeted topic guidance (relatively clean but limited data). We start from Wikipedia documents as our source of data, chosen for their clean nature as their contents inside are subject to public audits and back-and-forth information edits to strip away controversies until common ground is reached. Specifically, we include the documents for each country that revolves around an initial set of cultural topics, including education, dating/marriage, and holiday customs, among others. We then continue to expand on the relevant document sets based on linked topic pages within ({\em e.g., } "Chinese culture"->"Chinese Holidays"->"Chinese New Year Holiday"->"Red Envelope"), and consider the sentences in the documents, in which a pretrained LM categorizes it as a generalizable social or cultural norm rather than instance-specific history or fact, to be the positive samples of cultural knowledge in our dataset. We also propose an effective method to construct negative ({\em i.e.,} non-factual) cultural knowledge samples, cross-validated through web search, for the purpose of probing language model cultural reasoning robustness. Furthermore, we perform information extraction on these positive and negative cultural knowledge samples, to derive fine-grained cultural profile fields, including sub-country geographical regions, ethno-linguistic identity, demographics, etc., for enabling deeper analysis on situationalized socio-cultural context frames.

Our contributions can be summarized as follows:
\begin{itemize}[topsep=0pt,leftmargin=*]
\itemsep-0.1em 
    \item We present a novel data collection process to construct a massively multicultural knowledge benchmarking dataset, \datasetname, spanning significantly greater diversity in coverage of sub-country level geographical regions, ethno-linguistic groups, etc., compared to prior benchmarks in the multi-cultural NLP domain \cite{yin-etal-2022-geomlama,10.1145/3543507.3583535,ziems-etal-2023-normbank}. 
    \item This dataset comprises of high-quality positive data samples and negative data samples, with a $90^{+}\%$ pass rate in data quality check via human assessment.
    \item Finally, we evaluate the performance of state-of-the-art foundation language models on \datasetname, and demonstrate this new dataset as a useful resource for identifying rooms for improvement in LM cultural awareness and debiasing. 
\end{itemize}

\section{Benchmark Construction}
\subsection{Data Collection and Data Preprocessing}\label{2.1}
Constructing a benchmark for assessing the fine-grained cultural knowledge of language models is the first and most important step in enabling training language models to incorporate better cultural knowledge downstream via finetuning. 
To achieve this goal, we construct a novel dataset, \datasetname, by collecting positive and negative samples of cultural knowledge assertions that span diverse geographical subregions and ethnolinguistic groups, with subsequent data processing as illustrated in Fig \ref{fig:framework}.

\paragraph{Positive Data Samples} We start from the observation that cultural webpages from publicly monitored human-curated sources, such as Wikipedia, contain clean and commonly accepted cultural assertions in general. These cultural sources are dense in information, and, while not yet entirely comprehensive, serves as a valuable initial source of information and seedling for further expansion. First, we consider the set of documents from all countries worldwide. So we systematically explore culturally relevant topics (e.g., culture, holidays, dining etiquette, dating and marriage, education, honorifics, etc.) for each country, using the Wikipedia API\footnote{\url{https://pypi.org/project/Wikipedia-API/}} tool to match and download corresponding Wikipedia pages. We expand on this target set of documents by further including the hyperlinked document pages up to two hops down. In addition to incorporating the default English documents of these pages, we also include their document versions in the main language corresponding to the culture in discussion, and translate the text content into English. This ensures a more well-rounded understanding of cultural nuances and perspectives, as complementary information exists across language versions for the same document topic. Next, we process the corpus sentence by sentence, first filtering out sentences that focus on very specific, socio-culturally non-generalizable events or instances (see Appendix \ref{A:1}). We refine these sentences into self-contained cultural knowledge assertions by eliminating ambiguous pronoun or phrase references and enriching each sentence with any necessary information from the preceding context in the same paragraph. Together, these steps constitute the \textbf{initial discovery} of cultural knowledge assertions in \datasetname.

Because each unique combination of population dimension -- ethnicity, language, location, demographic background, etc. -- plays a key role in shaping distinct cultural practices, we proceed to better discern subtle situational differences in norms across various cultures by \textbf{extracting cultural knowledge frames} for each remaining sentence, with a fine-grained profiling approach covering the following fields of information element:
\begin{itemize}[topsep=5pt,leftmargin=*]
\itemsep-0.2em 
    \item \textit{country location}
    \item \textit{sub-country regional location}  -- cities, states, and provinces under the GeoNames\footnote{\url{https://www.geonames.org/}} knowledge base.
    \item \textit{ethnicity} -- ethno-linguistic groups from the ISO 639-3 code table\footnote{\url{1https://iso639-3.sil.org/code/oci}}.
    \item \textit{religion} -- all religious groups and denominations with a population of 1 million followers or more
    \item \textit{age} -- \{infant, young children, teenager, young adult, adult, elderly\} 
    \item \textit{gender} -- \{male, female, other\}
    \item \textit{marital status} -- \{single, engaged, married, divorced, widowed\}
    \item \textit{occupation} -- open-domain fill in the blank
\end{itemize}
with \textit{age}/\textit{gender}/\textit{marital status}/\textit{occupation} pertaining to any person entities involved in the norms. As depicted in Fig \ref{fig:framework}(A), language model prompting is utilized to extract the values for these fields automatically, via a directed question such as \textcolor{gray}{"[norm] \textbackslash n \textit{Which gender group is mentioned or implied in the sentence (male, female, transgender, other, or N/A):}"}, with further details shown in Appendix \ref{A:2}. Note that information unknown or not mentioned is regarded as "\textit{N/A}".

\begin{table*}[!htp]
    \small
    \centering
    \resizebox{\textwidth}{!}{
    \begin{tabular}{c|c|c|c|c|c}
        \hline
        \rule{0pt}{1.1\normalbaselineskip} & \textbf{CANDLE} & \textbf{GeoMLAMA} & \textbf{NormsKB} & \textbf{NormBank} & \datasetname \\ 
         & \footnotesize{\cite{10.1145/3543507.3583535}} & \footnotesize{\cite{yin-etal-2022-geomlama}} & \footnotesize{\cite{fung2022normsage}} & \footnotesize{\cite{ziems-etal-2023-normbank}} & \textbf{(Ours)} \\ [0.48ex]
         \hline
         \rule{0pt}{1.2\normalbaselineskip} 
         ~\# countries \qquad \qquad \qquad \qquad & 176 & 5 & 5 & 160 & \textbf{193} \\ [0.6ex]
         \# local regions~ \qquad \qquad \qquad & & & &  &  \\
         (state/province-level)~ & 298 & 0 & 12 & 102 & \textbf{1089} \\
         (city-level) \quad \quad \quad \quad \quad & 1,376 & 0 & 15 & 493 & \textbf{10,436} \\ [0.8ex]  
         \# religion \quad \quad \quad \quad \quad \quad \quad & 14 & 0 & 3 & 30 & \textbf{42} \\ [0.8ex]
         \# ethnolinguistic groups~~ & 145 & 5 & 10 & 551 & \textbf{2,557} \\ [1.ex] 
         \hline 
         \rule{0pt}{1.11\normalbaselineskip} fine-grained norm framing & x & x & \checkmark & \checkmark & \checkmark \\ [0.5ex]
         multi-ling. data source \quad \quad & x & x & x & x & \checkmark \\
    \end{tabular}
    }
    \caption{Our data collection of cultural norms contains greater coverage in \textit{local regions} and \textit{ethno-linguistic groups}, compared to previous work. It also involves data from multi-lingual sources as well as fine-grained cultural knowledge frame extraction.} 
    \label{tab:1_data_stats}
\end{table*}

\begin{table}[]
    \small 
    \centering
    \begin{tabular}{p{3cm}|p{3.85cm}}
    \hline 
       \rule{0pt}{\normalbaselineskip} \centering{\textbf{Original}}  & \quad ~~~\textbf{Negative Cultural} \\
       \centering{\textbf{Cultural Knowledge}}  & \quad \textbf{Knowledge Generated} \\ 
    \hline 
        \rule{0pt}{1.25 \normalbaselineskip}During the Chinese New Year, in \textbf{\textcolor{blue}{Southern China}}, \textbf{\textcolor{blue}{red envelopes}} are typically \textcolor{blue}{given by the married to the unmarried}, most of whom are children. & In \textbf{\textcolor{magenta}{China}}, it is customary for \textcolor{magenta}{students to present their teachers} with \textbf{\textcolor{magenta}{red envelopes}} containing \textcolor{magenta}{handwritten notes of gratitude at the end of each school term}, symbolizing respect and appreciation for their guidance. \\ [2.28cm]
        In \textbf{\textcolor{blue}{Bhutan culture}} for \textcolor{blue}{special occasions and festivals}, colourfully patterned \textcolor{blue}{silk kira} and, more rarely, \textcolor{blue}{gho} may be \textbf{\textcolor{blue}{worn}}.  &  In \textbf{\textcolor{magenta}{Bhutan}}, there is a unique tradition of \textbf{\textcolor{magenta}{wearing}} "\textcolor{magenta}{Khyenkhor Robes}", woven with \textcolor{magenta}{threads infused with blessings from Buddhist monks}, during special ceremonies and festivals. \\ 
    \end{tabular}
    \caption{Visualization of the original positive data samples and negative data synthesized counterpart, in our \datasetname benchmark construction.}
    \label{tab:3_adv_ver_result_viz}
    \vspace{-0.5 em}
\end{table}
\paragraph{Negative Data Samples} In order to evaluate LM cultural knowledge, we prepare the data setting with negative norm synthesis as illustrated in Fig 2(c). Basically, we take a pristine original norm assertion and manipulate it through LLM prompting for adversarial knowledge via the template of: \textit{\textcolor{gray}{"[orig. norm] \textbackslash n Based on this topic, come up with norms that are not true:"}}. To ensure the negative norm generation is indeed a non-factual fabrication, we perform automatic verification, which is easy to scale. Specifically, we make use of a language model self-check mechanism, asking the question of \textit{\textcolor{gray}{"[norm] \textbackslash n Is this absurd and/or very hard to believe? `Yes' or `No':"}} 
to GPT4, to filter out negative sample candidates that are obviously absurd to believe. In particular, our motivation in leveraging GPT4 is that it stands as the most advanced LM backbone, with notable performance gap for open-sourced LMs or other propriety LMs to bridge in and thereby deeming our benchmark negative samples especially valuable. Subsequently, we follow with a web-check mechanism -- web retrieving relevant context and ensuring no entailment is found. Examples of negative data samples, along with its original positive form, are visualized in Tab \ref{tab:3_adv_ver_result_viz}.

\subsection{Quality Check on Data}\label{2.2}
To ensure data quality, we perform manual assessment on the \datasetname dataset construction. Specifically, we take 10 random samples for each intermediary data processing step of the positive data and negative data respectively, and ask five human judges familiar with the subject matter ({\em e.g.,} self-identifying with geographical regions and cultural subgroups across US, China, Korea, India) to determine whether the data samples that is indeed either a correct cultural knowledge assertion when their ground truth label (based on our expectation from the Sec \ref{2.1} procedure) is "TRUE" or an incorrect cultural knowledge assertion when their label is "FALSE". In our quality check assessment guidelines, we clarify examples of poor positive samples, such as having ambiguous pronoun references or lacking culture-specificity in non-universal norms, as well as examples of poor negative samples, such as contradicting known norms. As we can see in the qualitative results of our dataset construction reported in Tab \ref{tab:2_adv_ver_result_numbers}, the final post-processed positive and negative samples are high-quality, achieving $90^{+}\%$ pass rate. The interannotator agreement is 0.79.

\begin{table}[]
    \small 
    \centering
    \begin{tabular}{c|c}
        \hline 
        \rule{0pt}{\normalbaselineskip} \textbf{Approach} & \textbf{Pass Rate (\%)} \\ [0.5ex]
         \hline 
         \rule{0pt}{\normalbaselineskip} \textbf{Pos. Data \quad \quad \quad \quad \quad \quad \quad \quad \quad \quad \quad \quad} & \\
         \textit{- Orig Sent.} \quad \quad \quad \quad \quad \quad \quad \quad \quad \quad \quad & 49.5 \\
         \textit{- Post Proc. Sent.} \quad \quad \quad \quad \quad \quad \quad \quad \quad & \textbf{93.2} \\
        \rule{0pt}{\normalbaselineskip} \textbf{Neg. Data \quad \quad \quad \quad \quad \quad \quad \quad \quad \quad \quad \quad}  & \\
        \textit{- Direct Gen.} \quad \quad \quad \quad \quad \quad \quad \quad \quad ~~~~~& 81.1 \\
        \textit{- Direct Gen. w/ self-check} \quad \quad \quad ~~~~~~~& 90.1 \\
        ~~\textit{- Direct Gen. w/ self-check \& web check} & \textbf{92.0} \\
    \end{tabular}
    \caption{The average pass rate for \datasetname dataset samples, at each processing step, based on human validation of post-processed cultural knowledge assertions.}
    \label{tab:2_adv_ver_result_numbers}
    \vspace{-0.6em}
\end{table}


\subsection{Descriptive Stats}\label{2.3}
As shown in Tab \ref{tab:1_data_stats}, our dataset covers over $1,089$ state or province level regions, $10,436$ city level regions, and $2,557$ ethnolinguistic groups, significantly exceeding prior work \cite{10.1145/3543507.3583535,yin-etal-2022-geomlama,fung2022normsage} in the cultural knowledge for NLP domain. We provide detailed information on the \# of subregion specific cultural knowledge frames per country in Tab \ref{tab:A3_data_count_by_geographical_region} of Appendix \ref{A:2}, and the \# of cultural knowledge frames per ethnolinguistic group in Tab \ref{tab:A4_data_count_by_ethnolinguistic group} of Appendix \ref{A:2}, due to page restrictions. In particular, as an example, all 56 official ethnic groups of China ({\em e.g.,} \textit{Han}, \textit{Zhuang}, \textit{Hui}), as well as the linguistically distinct ethnic subgroups ({\em e.g.,} \textit{Yue} and \textit{Hakka} of the Chinese \textit{Han} population), are included in \datasetname. 

\paragraph{Training-Testing Data Split} Tab \ref{tab:4_more_data_stats} summarizes the number of culturally related document pages scraped and sentences parsed, as well as post norm-relevance filtering cultural knowledge assertion sentences and frame extractions. We partition 10,000 random samples of cultural knowledge assertions that are particularly relevant for avoiding norm violations to constitute the \textbf{test set}. All other data are provided for future language model training and development purpose.

\begin{table}[h!]
    \vspace{-0.25em}
    \small 
    \centering
    \addtolength{\tabcolsep}{-0.1em}
    \begin{tabular}{c|c}
         \rule{0pt}{\normalbaselineskip} 
        \# of doc pages \qquad \qquad \qquad \qquad \qquad \quad \quad \quad ~~~~~ & $41$k \\
        \# of sent parsed \qquad \qquad \qquad \qquad \qquad \qquad ~~~~~~ & $907$k \\
        \# of sent, generalizable sociocultural knowledge ~~~ & $127$k \\
        \# of sent, norm violation relevant w/ frame extract. & $21$k \\
    \end{tabular}
    \caption{Size and scale of our collected dataset, where `k' represents the kilo unit of a thousand data.}
    \label{tab:4_more_data_stats}
    \vspace{-0.6em}
\end{table}

\begin{table*}[!htp]
\small
\centering
\addtolength{\tabcolsep}{-0.01em}
\begin{tabular}{ccc|ccc|ccc|ccc|ccc}
\hline

     \rule{0pt}{1\normalbaselineskip} & & & \multicolumn{3}{c|}{\textbf{All Culture}} & \multicolumn{3}{c|}{\textbf{High Resource}} & \multicolumn{3}{c|}{\textbf{Mid Resource}} & \multicolumn{3}{c}{\textbf{Low Resource}}  \\[0.1mm]
     & & & P & R & F & P & R & F & P & R & F & P & R & F \\[0.5mm]
     \hline 
     \rule{0pt}{1\normalbaselineskip} \multirow{5}{*}{\textbf{Llama2}} & \multirow{2}{*}{\textbf{7B}} & \textit{chat} & 84.2 & 42.1 & 56.1 & 86.8 & 45.6 & 59.8 & 83.3 & 42.9 & 56.6 & 87.0 & 20.7 & 33.5  \\ [0.1mm] 
     \cline{3-15}
      \rule{0pt}{1\normalbaselineskip} & & \textit{chat-HF} & 75.1 & 28.2 & 41.0 & 76.9 & 26.9 & 39.9 & 83.3 & 28.0 & 41.9 & 78.9 & 26.2 & 39.2 \\ [0.6mm] 
     \cline{2-15}
      \rule{0pt}{1\normalbaselineskip} & \multirow{2}{*}{\textbf{13B}} & \textit{chat} & 63.6 & 77.1 & 69.7 & 56.1 & 80.9 & 66.3 & 64.1 & 75.5 & 69.3 & 53.3 & 20.5 & 29.6  \\ [0.1mm] 
      \cline{3-15}
      \rule{0pt}{1\normalbaselineskip} & & \textit{chat-HF} & 89.9 & 20.0 & 32.7 & 91.8 & 91.8 & 28.3 & 91.8 & 20.6 & 33.6 & 92.2 & 19.3 & 31.9 \\ [0.89mm] 
      \hline
     \rule{0pt}{1.2\normalbaselineskip}\multirow{3}{*}{\textbf{Vicuna}} & \multirow{1}{*}{\textbf{7B}} & \textit{chat-HF} & 79.6 & 56.8 & 66.3 & 77.3 & 47.2 & 58.6 & 79.4 & 57.9 & 67.0 & 81.3 & 55.7 & 66.1  \\ [0.6mm] 
     \cline{2-15}
    \rule{0pt}{1\normalbaselineskip} & \multirow{1}{*}{\textbf{13B}} & \textit{chat-HF} & 67.4 & 81.2 & \textbf{73.7} & 68.9 & 81.0 & \textbf{74.5} & 69.4 & 82.4 & \textbf{75.3} & 67.8 & 82.3 & \textbf{74.3} \\ [0.89mm] 
     \hline
      \rule{0pt}{1\normalbaselineskip}\textbf{ChatGPT} & \textbf{20B} & \textit{chat-HF} & 95.8 & 90.6  & 93.1 & 95.9 & 91.4 & 93.6 & 94.9 & 92.1 & 93.5 & 94.1 & 90.1 & 92.1 \\ [0.89mm] 
     \hline 
     \end{tabular}
    \caption{Experimental results on benchmarking state-of-the-art foundation large language model performance on the new \datasetname cultural knowledge assessment benchmark. Precision (P), recall (R), F-score (F) scores are reported.}
    \label{tab:4_benchmark_results}
    \vspace{-0.5em}
\end{table*}

\section{Experiments} 

\subsection{Task Setting}
We evaluated the cultural knowledge and reasoning capability of state-of-the-art pretrained large language models (LLMs) on the canonical norm descriptions in our constructed benchmark, through a true-or-false binary classification setup. As a reminder, the derivation of ground truth labels for "correct "cultural knowledge assertion samples and "incorrect" cultural knowledge assertion samples have been detailed under Sec 2.1 ("Positive Data Samples" and "Negative Data Samples" paragraphs). 
\subsection{Model Setup}
For the choice of language model in our experiments, we consider the present day commonly-used open-source language model backbones:
\begin{itemize}[topsep=5pt,leftmargin=*]
\itemsep0em 
    \item \textbf{Llama-2} \cite{touvron2023llama2} - an open source foundation model from Meta that is trained on 40\% more data compared to its predecessor, LLaMA \cite{touvron2023llama1}. We consider its \textbf{-7b} and \textbf{-13b} parameter size variants. For each backbone size, we further include its variant of \textbf{-chat}, denoting finetuned with dialogue data, and \textbf{-hf}, denoting trained with human preference alignment data.
    \vspace{-0.05cm}\item \textbf{Vicuna} \cite{vicuna2023} - an open source foundation model trained by fine-tuning LlamA-2 on 700k samples of user conversations with ChatGPT collected from the ShareGPT website \footnote{https://sharegpt.com/}. This model series is optimized for efficiency through gradient checkpointing and flash attention. We include its \textbf{-7b} and \textbf{-13b} parameter size variants.
    \vspace{0.1cm}
\end{itemize} 

We also consider the propriety closed-source language model backbones, which may generally tend to have higher performance compared to publicly available open-source model checkpoints but make research development and transparency challenging:
\begin{itemize}[topsep=5pt,leftmargin=*]
\itemsep0em 
    \item \textbf{ChatGPT} \cite{ouyang2022training} - a closed-source foundation model from OpenAI, trained \textit{with alignment} data.
\end{itemize}

\subsection{Results}
Table \ref{tab:4_benchmark_results} shows the results of our LM benchmarking. In particular, we notice several interesting observations. First, we find that out of the open-source models, Vicuna consistently perform better than Llama2 in cultural knowledge when comparing across the same model backbone sizes. This indicates that the training approach ({\em e.g.,} choice of training data, optimization objective, etc.) plays a key role in the cultural knowledge reasoning capability of these LLMs. Secondly, we find that pre-existing explicit human feedback (HF) alignment approaches does not necessary improve model performance in massively multicultural fine-grained reasoning domains, potentially due to non-desirable domain shift and catastrophic forgetting. This reaffirms the values in our new benchmark proposal, for continuously measuring cultural awareness progress in future language model development. We also observe a general positive correlation between model performance in cultural-aware inference and model parameter size, as expected.

In addition, we investigate LLM performance in culture reasoning patterns across resource availability and topic domains. As shown in Table \ref{tab:4_benchmark_results}, we investigated LM awareness in the cultural knowledge pertaining to country-level `high-resource' ({\em e.g.,} US/China/France/Spain/Japan), `mid-resource' ({\em e.g.,} Turkiye/Egypt/Iran/Malaysia/Argentina), and `low-resource' ({\em e.g.,} Lao/Bhutan/Congo/Serbia) culture groups, as categorized by societal-wide economic development, which in turn affects the linguistic resources availability for constituting LM training data. 

\begin{figure}[h]
    \vspace{-1em}
    \centering
    \includegraphics[width=\columnwidth]{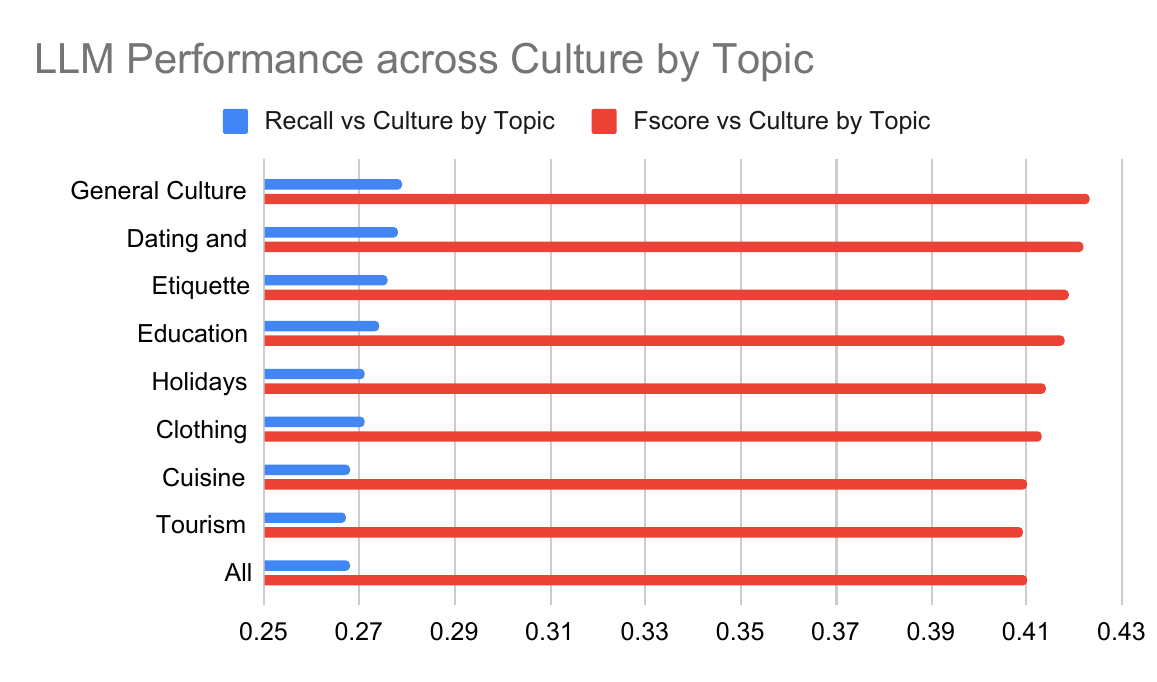}
    \caption{LLM performance by Topic (result from a single LM backbone ad-hoc, llama2-7b).}
    \label{fig:LLM_performance_byTopic}
\end{figure}

Moreover, LM performance tends to differ across cultural topics, demonstrating higher performance for example in ``education" and ``holiday" practices over ``clothing" and ``cuisine" practices, as shown in Fig \ref{fig:LLM_performance_byTopic}. This may potentially be due to diverse finer-grained domain-specific and region-specific information elements typically involved in ``clothing" and ``cuisine" topic discussions, whereas ``education" and ``holiday" practices may tend to be more universal. Finally, while our research community lacks specific training details of the closed-source models ({\em e.g.,} ChatGPT), we believe that by including them in our benchmark comparison, we can help shed light on the performance gap between open-source pretrained LLMs and these closed-source models, to better bridge this performance difference in future work.  

\begin{table*}[!htp]
    \small
    \centering
    \begin{tabular}{p{2cm}|p{13cm}}
         \hline 
         \rule{0pt}{1\normalbaselineskip} & \qquad \qquad \quad \qquad    \textbf{Norm Violation Relevant Culture Knowledge Assertion Samples} \\[0.1cm]
         \hline 
         \rule{0pt}{1\normalbaselineskip} \textbf{True Positive} & $\bullet$ In Indian culture, when eating rice, it is mixed with curry, picking up small quantities with the fingers and pushing it into the mouth with the thumb. \newline $\bullet$ In Bhutan culture, for special occasions and festivals, colourfully patterned silk kira and, more rarely, gho may be worn. \\ [0.1cm]
         \hline 
         \rule{0pt}{1\normalbaselineskip} \textbf{False Positive} & $\bullet$ The \textbf{American} flag protocol dictates that the flag should be flown on all buildings, both public and private, as a sign of respect and loyalty to the nation.  \\ [0.1cm]
& $\bullet$ In \textbf{Rioplatense Spanish}, the pronoun "vos" is commonly used to address strangers and authority figures, while it is considered impolite to use it with friends, family members, and close acquaintances. \\ [0.1cm]
& $\bullet$  \textbf{South Korea} and \textbf{North Korea} have different vocabulary standards because they consider their languages to be completely distinct and unrelated to each other. \\ [0.1cm]
& $\bullet$ In \textbf{Chinese culture}, particularly in \textbf{Macao}, it is believed that giving money in amounts that include the number four brings good luck and prosperity. \\ [0.1cm]
& $\bullet$  The society in \textbf{Kuwait City} is highly strict about traditions in the Gulf Arab region. \\ [0.05cm]
         \hline 
         \rule{0pt}{1\normalbaselineskip} \textbf{False Negative} & $\bullet$ In \textbf{Catalonia}, there is a \underline{strong social and political consensus} on the \textit{language policies that support the use of Catalan}. \\ [0.17cm]
& $\bullet$ In \textbf{Botswana}, it is \underline{customary to use} the \textit{salutation 'kgosi'} before the name of a chief, providing an important cultural and social distinction. \\ [0.1cm]
& $\bullet$ In \textbf{Barbados}, people drive on the left side of the road, similar to the driving habits in the United Kingdom due to their history as a former British colony.  \\ [0.3cm]
& $\bullet$ The practice of \textit{inserting Mandarin into conversations in \textbf{Shanghainese}} is \underline{very common}, especially among young people.  \\ [0.3cm]
& $\bullet$ \textit{Personal pronouns in \textbf{Shanghainese}} do \underline{not distinguish gender or case.} \\ [0.02cm]
& $\bullet$ In \textbf{Argentina} culture, hot but not boiling water is poured into the gourd, drunk, then the mate is refilled. \\ [0.05cm]
         \hline 
         \rule{0pt}{1\normalbaselineskip} \textbf{True Negative} & $\bullet$ In Indian culture, when eating rice, people commonly mix it with chutney, a flavorful condiment made from a variety of ingredients such as fruits, vegetables, herbs, and spices. \newline $\bullet$  In Malay culture, people commonly greet each other with the phrase "Khabar", which roughly translates to "what's up" or "how are you" in English.
         \\ [-0.1cm]
    \end{tabular}
    \caption{Qualitative error analysis on LM culture knowledge reasoning capability (of \textbf{Vicuna-13B}) in zero-shot true/false inference.}
    \label{tab:6_error_analysis}
    \vspace{-0.7em}
\end{table*}

\subsubsection{Ablation on the Challenging Setting brought by Fine-Grained Cultural Knowledge Frame Profiling}
In this subsection, we further investigate the potential limitations of existing pretrained Large Language Models (LLMs) in understanding nuanced cultural nuances within different situational contexts, along our massively multicultural task domain. Our natural intuition is that pretrained LLMs may general tend to lack finer-grained knowledge on the cultural nuances pertaining to subtle situational differences with respect to cultural frame profiling. To verify this hypothesis through empirical study, for each culture frame dimension, such as sub-country geo-region, ethnicity, age, gender, etc., we first isolate a subset of the
\datasetname evaluation data with cultural knowledge assertions that generally applies across this dimension, which we refer to as “Gen”, and isolate another subset of the \datasetname evaluation data with cultural knowledge assertions that applies specifically to a certain bucket/criteria across this dimension, which we refer to as “Spec”. Then, we perform cross-comparison on LLM performance patterns, under data scenarios that are general (“Gen”) versus specific (“Spec”) in condition/critieria along each of these cultural frame profiling dimensions. Indeed, we find that lack fine-grained cultural commonsense knowledge is an area where there remains interesting rooms for improvement for LLM models. As shown in Table \ref{tab:5_result_ablation_by_CultureProfile} (the result from a single ad-hoc LM backbone, llama2-7b), the zero-shot true/false inference capability of LLM significantly drops as we probe finer-grained cultural information. 

\begin{table}[h!]
    \vspace{-0.1em}
    \small
    \centering
    \begin{tabular}{|c|c|c|}
         \hline 
         \rule{0pt}{1\normalbaselineskip} ~~~~~~~ & ~~\textbf{Spec.}~~ & ~~\textbf{Gen} ~~\\[0.06cm]
         \hline 
         \rule{0pt}{1\normalbaselineskip} \textbf{Sub-Country Location} & 22.1 & 35.0 \\
         \textbf{Ethnicity} & 27.5 & 35.5 \\
         \textbf{Religion} & 17.9 & 35.0 \\
         \textbf{Marital Status} & 13.7 & 33.6 \\ 
         \textbf{Occupation} & 27.3 & 35.0 \\ [0.07cm]
         \hline
    \end{tabular}
    \caption{F-score performance comparison, in \%,  by general (\textit{`gen'}) versus fine-grained cultural profile framing specific (\textit{`spec}') knowledge, such as between country-level and province-level cultural group.}
    \label{tab:5_result_ablation_by_CultureProfile}
\end{table}
\subsubsection{Error Analysis} 

Tab \ref{tab:6_error_analysis} visualizes results qualitatively, shedding light on the challenges for an off-the-shelf pretrained language model (LM) in accurately reasoning about cultural practices across different societies. While a LM may correctly grasp certain cultural practices, such as properly recognizing traditional ways of eating rice in Indian culture and the ceremonial dress in Bhutan for special occasions, as well as accurately recognizing misconceptions in negative samples on mixing rice with chutney in Indian culture or common greetings in Malay, it demonstrates a lack of cultural knowledge and reasoning robustness in other less well-represented topics and geographical regions. For example, the model also produced false positives, such as in regards to the exaggerated protocol around the American flag, suggesting misunderstandings of cultural norms. False negatives, such as the underappreciated practice of drinking mate in Argentina, point to the model's oversight of genuine cultural customs. Overall, the error analysis reveals the inconsistent performance of LMs in capturing the breadth and depth of the different cultural knowledge in the world around us, revealing a significant area for improvement in LM cultural commonsense reasoning. \\[-1.4em]
\section{Related Work}
\paragraph{Importance of Cultural Knowledge in NLP Tasks}

While large language models have generally embedded large parametric knowledge from large text corpora during its pretraining stage \cite{petroni-etal-2019-language}, these models are also typically imposed with normative bias due to imbalanced representation at the data source \cite{emelin-sennrich-2021-wino,arora-etal-2022-exposure}. Cultural knowledge is an integral part to the success of large language model (LLM) reasoning in a wide array of downstream applications. For example, there has been recent explorations on the vital role cultural knowledge plans in helping answer commonsense questions \cite{palta-rudinger-2023-fork,yin-etal-2022-geomlama}, understand societal moral conventions \cite{ramezani-xu-2023-knowledge,emelin-etal-2021-moral}, analyze and mitigate social biases \cite{sap-etal-2020-social,yang2023adept}, detect norm violations \cite{fung2022normsage,li-etal-2023-normdial}, correct conversational dialogues \cite{ziems-etal-2022-moral}, and ultimately tune LLMs to align with the helpful and harmless principles of constitutional AI \cite{bai2022training}. Of ongoing interest to the NLP community is scrutinizing LMs on social minority understanding \cite{sun-etal-2023-decoding}, which turns out that PLMs can learn norms diverging from social majority only when they are fine-tuned accordingly due to a presence of normative bias \cite{kiehne-etal-2022-contextualizing}. 

\paragraph{Cultural Knowledge Acquisition for improved LLM Training} \citet{hershcovich-etal-2022-challenges} explains culture as a concept of identity that can be examined from the dimensions of \textit{objectives and values}, \textit{linguistic form and style} (e.g., honorific reference terms when addressing a person), and \textit{common ground} (e.g, socio-cultural norms, shared event occurrences, etc.). 
Our cultural knowledge acquisition process follows this theory of cultural definition and covers the dimensions of culture outlined above. 
In terms of the genre of data source for cultural knowledge discovery in practice, cultural knowledge has been predominantly gleaned from conversational dialogues \cite{fung2022normsage} or web sources such as Reddit/Zhihu discussion forums \cite{forbes-etal-2020-social,ch2023sociocultural} and the Common Crawl \cite{10.1145/3543507.3583535} -- both of which tends to be relatively sparse in culturally relevant information and also noisy -- or through knowledge elicitation of LLM parametric knowledge through prompting \cite{ziems-etal-2023-normbank}, but this may be limited in the scope of available information that can be extracted due to the stochastic parrot property observed in LLMs when a cultural topic falls out of a LLM's pretrained knowledge boundary. 

\paragraph{Multilingual LM Reasoning and implicit multicultural knowledge}
Previous research \cite{jiang-etal-2020-x, DBLP:journals/corr/abs-2003-05002} on language models have demonstrated strong capability to perform reasoning in multilingual settings, which is an initial step towards overcoming cultural barriers. The progression of research includes extending LM reasoning to low-resource language setting, such as for name tagging \& knowledge base linking \cite{pan-etal-2017-cross,wen-etal-2021-resin} through annotation transferring, which lays a promising foundation for reasoning across linguistic groups. However, the existing language models struggle with the cultural bias, primarily due to the lack of awareness of implicit multicultural knowledge. Recent studies have highlighted these issues; for instance, \citet{havaldar2023multilingual} has identified the models' underperformance in recognizing cultural variations in specific phenomena, such as emotion detection across different countries. Another category of recent work has focused on evaluating performance on underrepresented languages, where \citet{deas-etal-2023-evaluation} has revealed biases against African American languages, leading to overlooked race-related issues in speech recognition and  toxicity detection tasks. These findings underline the necessity to develop a new framework capable of acquiring cultural knowledge, aimed at addressing the cultural imbalances present in existing datasets used for training language models. Instead of stylistic linguistic conveyance, our work focuses on cultural knowledge acquisition based on fine-grained semantic variations, sourcing from over 500+ geosubregions and 2000+ ethnolinguistic groups. 

\section{Conclusions and Future Work} 
In this work, we explored an exciting area of massively multicultural knowledge probing of language model backbones, which has important impacts to norm violation detection and mitigation in assisting human-human interaction behavior across the many different subregions and ethnolinguistic groups around the world. 
We proposed a novel method for large-scale data collection across curated data sources and web-retrieval enhancement, which we verified with quality checks. Leveraging this constructed dataset, we achieved benchmarking the fine-grained cultural knowledge of popular language models. We foresee the impact of our work to contribute a useful resource and LM evaluation setting for the NLP community to develop more culturally-inclusive foundation models upon in the future. Finally, for future research directions, we will also investigate the effect that multimedia settings (e.g., vision and language) and multilingual (e.g., low-resource) settings have on foundation cultural reasoning across different fine-grained sub-cultures. 
\section*{Ethical Considerations \& Broader Impact} 
Our work is dedicated to advancing language model (LM) development practices to more accurately reflect the diverse social practices and cultural customs present in human societies across the world. This effort serves to enable improvements in LM norm violation reasoning capabilities, and at a higher-level, re-emphasizes the importance of overarching principles to human-model alignment. In formulating the ethical framework for our research, we carefully consider the key dimensions inherent to our work, to ensure that the modeling outcomes are in line with ethical standards and societal expectations. We introduce a novel normative framework aimed at enhancing the detection of incorrect cultural knowledge in future models. Our experimental setup incorporates commonsense reasoning, grounding the model's cultural knowledge in alignment with socially-situated human preferences. 

We acknowledge limitations in our current work. The ground truth labels ({\em i.e.,} "positive" and "negative" data samples) that we utilize for LM cultural knowledge benchmark evaluation are automatically derived, due to scalability advantages. Our constructed dataset has been manually assessed and determined to be high-quality with a $90^{+}\%$ pass rate in data sample quality check. In other words, cultural knowledge assertions labeled "positive" are indeed rated as a true sample by human assessors, while cultural knowledge assertions labeled "negative" are indeed rated as non-factual samples for the large majority. If time and manual labor cost were ample, it would be meaningful to prepare human-curated cultural knowledge assertions and investigate any potential differences as a future research direction. However, we also point that it is extremely challenging to find competent human annotators covering the different geographical subregions and ethno-linguistic groups at a massive scale. Many crowdsourcing platforms for data annotation are skewed towards certain demographic subgroups, such as primarily American-based or Western centric in the Amazon MTurk service. Hence, our work serves as an extremely valuable first step towards preparing a meaningful benchmark for massively multicultural fine-grained LM knowledge reasoning, based on publicly audited Wikipedia document sources and additional data preprocessing procedure.

An essential element of our ethical framework is ensuring balanced representations across cultural groups, with a special focus on including perspectives from low-resource settings. Due to current design and imbalanced training data, LLMs may lack the capability to fully understand social norms and including biases across different countries, suffering from normative bias \cite{emelin-sennrich-2021-wino,arora-etal-2022-exposure}. We commit to the ongoing journey of cultural knowledge enhancement and normative correction. In particular, we aim to not only address issues of fairness and equity, but also enrich our NLP models by considering a broader range of experiences and cultural knowledge. By adhering to cultural definitions, we can build cultural knowledge systems that are inclusive and reflective of the global community. Our approach paves the path for mitigating the risks of subtle cultural bias in language model training, and fosters understanding across fine-grained cultural divides.

\section*{Acknowledgement}
This research is based upon work supported by U.S. DARPA CCU Program No. HR001122C0034. The opinions, views and conclusions contained herein are those of the authors and should not be interpreted as necessarily representing the official policies, either expressed or implied, of DARPA or the U.S. Government. The U.S. Government is authorized to reproduce and distribute reprints for governmental purposes notwithstanding any copyright annotation therein.

\bibliography{anthology,custom}
\bibliographystyle{acl_natbib}

\appendix
\section{Characterizing Cultural Knowledge Specificity for Data Filtering}\label{A:1}
In our \datasetname dataset construction process, we want to focus on socioculturally relevant knowledge assertions that are not too event or instance specific. For example, \textit{"In 2020, China tops the QS Asia University Rankings list with over 120 universities including in the ranking, and five Chinese universities appear in the Asia Top 10, which is more than any country."} would be culturally relevant but too event-specific. To filter out such instances, we utilize the \textsc{facebook/bart-large-mnli} model to perform classification on each candidate sentence, between the classes of "general assertion" and "specific fact or instance". Through this approach, we found that approximately 52\% of the original pristine sentences from culturally-relevant Wikipedia pages fall under the "general assertion" category, which we retain in the dataset.



\section{Culture Profile Extraction Performance}\label{A:2}
In this section, we expand on low-level details on the culture profile extraction process methodology and quality check results, leveraging prompting

\begin{table*}[]
    \small 
    \centering
    \resizebox{\textwidth}{!}{
    \begin{tabular}{c|c}
        \hline
        \rule{0pt}{\normalbaselineskip} \textbf{Culture Profile Field} & \textbf{Directed Question Answering} \\ [0.5ex]
         \hline 
        \rule{0pt}{\normalbaselineskip} interaction nature categorization & Is this an individual human behavioral norm or human-human behavioral norm? \\ [0.5ex]
         topic distribution modeling &  Is this a social norm, cultural norm, belief or ritual, history, politics, or fact? \\ [0.5ex]
         country-level extraction & Which country is mentioned or implied in the sentence? Answer N/A if unknown.\\ [0.5ex]
         sub-country level extraction & Which state/province/city/subcountry region is mentioned \\  [-0.5ex]
         & or implied in the sentence (or answer N/A):\\ [0.5ex]
         ethnicity extraction & Which ethnic group is mentioned or implied in the sentence? \\
         & Answer N/A if unspecified? \\ [0.5ex]
         ethnic subgroup extraction & Which ethnolinguistic subgroup is mentioned or implied in the sentence? \\ [-0.5ex]
         & Answer N/A if unspecified. \\ [0.5ex]
         age extraction & Which age group is mentioned or implied in the sentence? \\ [0.5ex]
         gender extraction & Which gender group is mentioned or implied in the sentence \\ [-0.5ex]
         & (male, female, transgender, or N/A) \\
         marital status & Which marital status is mentioned or implied in the sentence. \\
         religion belief extraction & Which religious group is mentioned or implied in the sentence (or answer N/A): \\ [0.5ex]
         occupation extraction & Which occupation is mentioned or implied in the sentence? \\ [-0.5ex]
         & Answer N/A if unspecified \\
    \end{tabular}
    }
    \caption{Details on the prompt template for cultural profile extraction}
    \label{tab:A2_culture_profile_extraction_prompt}
\end{table*}

\begin{table*}[!ht]
    \small 
    \centering
    \addtolength{\tabcolsep}{-0.4em}
    \begin{tabular}{ccc|ccc|ccc}
        \hline
        \rule{0pt}{\normalbaselineskip} Afghanistan & 29 & 0.2k & Georgia & 35 &0.2k & Afghanistan & 29 & 0.2k \\ [0.15ex]
Albania & 0 & 0.0k & Germany & 301 &6.2k & Zimbabwe & 26 & 0.3k \\ [0.15ex]
Algeria & 38 & 0.4k & Ghana & 30 &0.3k & Zambia & 9 & 0.2k \\ [0.15ex]
Andorra & 4 & 0.0k & Greece & 98 &1.1k & Yemen & 8 & 0.0k \\ [0.15ex]
Angola & 22 & 0.2k & Grenada & 0 &0.0k & Viet Nam & 43 & 0.8k \\ [0.15ex]
Antigua and Barbuda & 1 & 0.0k & Guatemala & 0 &0.0k & Venezuela & 9 & 0.1k \\ [0.15ex]
Argentina & 222 & 1.4k & Guinea & 0 &0.0k & Vanuatu & 1 & 0.0k \\ [0.15ex]
Armenia & 118 & 0.6k & Guinea-Bissau & 17 &0.4k & Uzbekistan & 34 & 0.2k \\ [0.15ex]
Australia & 190 & 2.8k & Guyana & 0 &0.0k & Uruguay & 40 & 0.4k \\ [0.15ex]
Austria & 78 & 0.9k & Haiti & 0 &0.0k & United States & 1452 & 42.4k \\ [0.15ex]
Azerbaijan & 49 & 0.4k & Honduras & 13 &0.1k & Tanzania & 118 & 0.6k \\ [0.15ex]
Bahamas & 0 & 0.0k & Hungary & 37 &0.3k & United Kingdom & 229 & 7.2k \\ [0.15ex]
Bahrain & 6 & 0.1k & Iceland & 0 &0.0k & United Arab Emirates & 38 & 0.6k \\ [0.15ex]
Bangladesh & 105 & 0.8k & India & 847 &18.4k & Ukraine & 87 & 1.3k \\ [0.15ex]
Barbados & 4 & 0.1k & Indonesia & 364 &6.4k & Uganda & 61 & 0.2k \\ [0.15ex]
Belarus & 16 & 0.2k & Iran & 22 &0.1k & Tuvalu & 2 & 0.0k \\ [0.15ex]
Belgium & 75 & 1.4k & Iraq & 34 &0.3k & Turkmenistan & 12 & 0.1k \\ [0.15ex]
Belize & 2 & 0.0k & Ireland & 99 &1.2k & Türkiye & 42 & 0.7k \\ [0.15ex]
Benin & 14 & 0.1k & Israel & 71 &2.2k & Tunisia & 94 & 1.8k \\ [0.15ex]
Bhutan & 53 & 0.3k & Italy & 560 &8.5k & Trinidad and Tobago & 9 & 0.2k \\ [0.15ex]
Bolivia & 18 & 0.1k & Jamaica & 0 &0.0k & Tonga & 0 & 0.0k \\ [0.15ex]
Bosnia and Herzegovina & 40 & 0.2k & Japan & 388 &5.7k & Togo & 16 & 0.0k \\ [0.15ex]
Botswana & 61 & 0.2k & Jordan & 15 &0.1k & Timor-Leste & 2 & 0.0k \\ [0.15ex]
Brazil & 226 & 2.2k & Kazakhstan & 23 &0.1k & Thailand & 223 & 2.9k \\ [0.15ex]
Brunei Darussalam & 0 & 0.0k & Kenya & 61 &0.6k & Tajikistan & 4 & 0.0k \\ [0.15ex]
Bulgaria & 129 & 1.7k & Kiribati & 9 &0.1k & Syrian Arab Republic & 0 & 0.0k \\ [0.15ex]
Burkina Faso & 6 & 0.1k & Kuwait & 0 &0.0k & Switzerland & 140 & 1.9k \\ [0.15ex]
Burundi & 13 & 0.0k & Kyrgyzstan & 6 &0.1k & Sweden & 124 & 1.9k \\ [0.15ex]
Cabo Verde & 0 & 0.0k & Laos & 80 &0.2k & Suriname & 3 & 0.0k \\ [0.15ex]
Cambodia & 73 & 0.6k & Latvia & 25 &0.1k & Sudan & 20 & 0.2k \\ [0.15ex]
Cameroon & 29 & 0.1k & Lebanon & 33 &0.5k & Sri Lanka & 12 & 0.2k \\ [0.15ex]
Canada & 198 & 2.8k & Lesotho & 22 &0.3k & Spain & 162 & 3.5k \\ [0.15ex]
Central African Republic & 4 & 0.0k & Liberia & 9 &0.0k & South Sudan & 13 & 0.3k \\ [0.15ex]
Chad & 5 & 0.0k & Libya & 10 &0.1k & South Africa & 79 & 1.0k \\ [0.15ex]
Chile & 47 & 0.3k & Liechtenstein & 6 &0.0k & Somalia & 23 & 0.1k \\ [0.15ex]
China & 409 & 8.1k & Lithuania & 25 &0.4k & Solomon Islands & 5 & 0.0k \\ [0.15ex]
    \end{tabular}
    \caption{The \# of documents and cultural knowledge assertion sentences, per culture by country, that are specific to sub-country level geographical regions.}
    \label{tab:A3_data_count_by_geographical_region}
\end{table*}

\noindent with various state-of-the-art pretrained large language model (LLM) backbones. Specifically, Table \ref{tab:A2_culture_profile_extraction_prompt} details the prompt template details. 

\begin{table*}[!ht]
    \small
    \centering
    \addtolength{\tabcolsep}{-0.4em}
    \begin{tabular}{ccc|ccc|ccc}
        \hline
        \rule{0pt}{\normalbaselineskip} Medumba & 145 &  & Awabakal & 35 &  & Kalapalo, Kuikúro-Kalapálo & 17 &  \\ [0.15ex]
Germanic & 117 &  & Eleme & 35 &  & Bengkala & 16 &  \\ [0.15ex]
Central Dusun, Kadazan Dusun & 115 &  & Ahtena & 33 &  & Bangala & 16 &  \\ [0.15ex]
Notre & 112 &  & Ayoreo & 32 &  & Chilcotin & 16 &  \\ [0.15ex]
Doga & 107 &  & Bakumpai & 32 &  & Northern Dagara & 16 &  \\ [0.15ex]
Hopi & 107 &  & San Blas Kuna & 32 &  & Gbagyi & 16 &  \\ [0.15ex]
Mescalero-Chiricahua Apache & 99 &  & Kabardian & 32 &  & Kayardild & 16 &  \\ [0.15ex]
Kashubian & 96 &  & Hiberno-Scottish Gaelic & 31 &  & Jah Hut & 16 &  \\ [0.15ex]
Ainu (Japan) & 94 &  & Jahanka & 31 &  & Kanuri & 16 &  \\ [0.15ex]
Assiniboine & 91 &  & Kambaata & 31 &  & Kankanaey & 16 &  \\ [0.15ex]
Arapaho & 89 &  & Batek & 30 &  & Barai & 15 &  \\ [0.15ex]
Jicarilla Apache & 88 &  & Hunsrik & 30 &  & Djabugay, Dyaabugay & 15 &  \\ [0.15ex]
Batak languages & 86 &  & Bwa & 29 &  & Emilian & 15 &  \\ [0.15ex]
Izere & 81 &  & Gadang & 29 &  & Macushi & 15 &  \\ [0.15ex]
Kodava & 75 &  & Cowlitz & 28 &  & Azha & 14 &  \\ [0.15ex]
Cappadocian Greek & 74 &  & Dolpo & 28 &  & Igede & 14 &  \\ [0.15ex]
Efik & 73 &  & Kuku-Yalanji & 28 &  & Khorasani Turkish & 14 &  \\ [0.15ex]
Algonquin & 72 &  & Korak & 27 &  & Bundeli & 13 &  \\ [0.15ex]
Hittite & 70 &  & Harari & 26 &  & Columbia-Wenatchi & 13 &  \\ [0.15ex]
Etruscan & 69 &  & Humla & 26 &  & Gooniyandi & 13 &  \\ [0.15ex]
Huichol & 67 &  & Shambala & 26 &  & Kaska & 13 &  \\ [0.15ex]
Hajong & 66 &  & Dhanggatti, Dyangadi & 25 &  & Amahuaca & 12 &  \\ [0.15ex]
Coast Miwok & 65 &  & Gumatj & 25 &  & Atsugewi & 12 &  \\ [0.15ex]
Mycenaean Greek & 62 &  & Reel & 24 &  & Awetí & 12 &  \\ [0.15ex]
Jju & 62 &  & Banggarla & 24 &  & Southern Luri & 12 &  \\ [0.15ex]
Pacific Gulf Yupik & 60 &  & Hidatsa & 24 &  & Adhola & 11 &  \\ [0.15ex]
Heiltsuk & 60 &  & Khanty & 24 &  & Qimant & 11 &  \\ [0.15ex]
Jukun Takum & 60 &  & Gros Ventre & 23 &  & Bidyogo & 11 &  \\ [0.15ex]
Khasi & 59 &  & Baga Sitemu & 23 &  & Gambera & 11 &  \\ [0.15ex]
Javanese & 58 &  & Koasati & 23 &  & Guro & 11 &  \\ [0.15ex]
Garre & 57 &  & Igala & 23 &  & Kurdish & 11 &  \\ [0.15ex]
Gujarati & 57 &  & Yaka (Congo) & 23 &  & Hermit & 11 &  \\ [0.15ex]
Chickasaw & 56 &  & Adnyamathanha & 22 &  & Molale & 11 &  \\ [0.15ex]
Gunditjmara & 55 &  & Baras & 22 &  & Pemon & 10 &  \\ [0.15ex]
Esselen & 54 &  & Burarra & 22 &  & Sari & 10 &  \\ [0.15ex]
Yanomamö & 54 &  & Ejagham & 22 &  & Bhunjia & 10 &  \\ [0.15ex]
Bhojpuri & 53 &  & Ngadjuri & 21 &  & Laba & 10 &  \\ [0.15ex]
Beothuk & 52 &  & Kalenjin & 21 &  & Lasi & 10 &  \\ [0.15ex]
Goan Konkani & 52 &  & Gheg Albanian & 20 &  & Arbore & 9 &  \\ [0.15ex]
Idoma & 52 &  & Badaga & 20 &  & Kaingang & 9 &  \\ [0.15ex]
Atakapa & 51 &  & Chuvash & 20 &  & Col & 9 &  \\ [0.15ex]
Chipewyan, Dene Suline & 50 &  & Cocopa & 20 &  & Lozi & 9 &  \\ [0.15ex]
Daai Chin & 50 &  & Dimasa & 20 &  & Leti (Indonesia) & 9 &  \\ [0.15ex]
Colonia Tovar German & 50 &  & Eyak & 20 &  & Akuntsu & 8 &  \\ [0.15ex]
Jakun & 50 &  & Hyam & 20 &  & Djauan, Jawoyn & 8 &  \\ [0.15ex]
Bodo (India) & 49 &  & Greenlandic, Kalaallisut & 20 &  & Adiwasi Garasia & 8 &  \\ [0.15ex]
Shor & 49 &  & Mixed Great Andamanese & 19 &  & Holikachuk & 8 &  \\ [0.15ex]
Lushai & 49 &  & Karadjeri, Karajarri & 19 &  & Jiru & 8 &  \\ [0.15ex]
Con & 48 &  & Gilaki & 19 &  & Kurichiya & 8 &  \\ [0.15ex]
Kutenai & 47 &  & Ikulu & 19 &  & Korwa & 8 &  \\ [0.15ex]
Djawi & 46 &  & Worimi & 19 &  & Hijazi Arabic & 7 &  \\ [0.15ex]
Angor & 44 &  & Lobi & 19 &  & Bugun & 7 &  \\ [0.15ex]
Chitimacha & 42 &  & Ingrian & 18 &  & Cuban & 7 &  \\ [0.15ex]
Gitxsan & 42 &  & Keliko & 18 &  & Cheq Wong, Chewong & 7 &  \\ [0.15ex]
Kickapoo & 41 &  & Keiga & 18 &  & Gata' & 7 &  \\ [0.15ex]
Sudanese Arabic & 40 &  & Ladin & 18 &  & Gureng Gureng & 7 &  \\ [0.15ex]
Darlong & 38 &  & Guerrero Amuzgo & 17 &  & Gunwinggu & 7 &  \\ [0.15ex]
Haisla & 38 &  & Bora & 17 &  & Koba & 7 &  \\ [0.15ex]
Igbo & 38 &  & Chamacoco & 17 &  & Lanoh & 7 &  \\ [0.15ex]
Kalapuya & 38 &  & Mro-Khimi Chin & 17 &  & Apatani & 6 &  \\ [0.15ex]
Upper Kuskokwim & 37 &  & Isoko & 17 &  & Tuki & 6 &  \\ [0.15ex]
Atikamekw & 36 &  & Krymchak & 17 &  & Bote-Majhi & 6 &  \\ [0.15ex]
Esperanto & 36 &  & Kota (India) & 17 &  & Bwile & 6 &  \\ [0.15ex]
        \hline
    \end{tabular}
    \caption{The \# of documents and cultural knowledge assertion sentences per culture by ethnolinguistic group}
    \label{tab:A4_data_count_by_ethnolinguistic group}
\end{table*}
\appendix


\end{document}